\documentclass[lettersize,journal]{IEEEtran}
\usepackage{amsmath,amsfonts}
\usepackage{algorithmic}
\usepackage{algorithm}
\usepackage{array}
\usepackage[caption=false,font=normalsize,labelfont=sf,textfont=sf]{subfig}
\usepackage{textcomp}
\usepackage{stfloats}
\usepackage{url}
\usepackage{verbatim}
\usepackage{booktabs}
\usepackage{graphicx}
\usepackage{cite}
\usepackage{multirow}
\usepackage{float}
\begin{document}

\title{3D Hand-Eye Calibration for Collaborative Robot Arm: Look at Robot Base Once}

\author{
    Leihui Li\textsuperscript{\textdagger,1}, Lixuepiao Wan\textsuperscript{\textdagger,1}, Volker Krueger\textsuperscript{2} and Xuping Zhang\textsuperscript{*,1}\thanks{\textsuperscript{\textdagger}These authors contributed equally to this work.} \thanks{\textsuperscript{1}Department of Mechanical and Production Engineering, Aarhus University, Aarhus, Denmark\par \textsuperscript{2}Department of Computer Science, Lund University, Lund, Sweden\par
    $^*$Corresponding author: Xuping Zhang, Email: xuzh@mpe.au.dk.}
}

\markboth{\;}%  
{\;}



\maketitle

\begin{abstract}
Hand-eye calibration is a common problem in the field of collaborative robotics, involving the determination of the transformation matrix between the visual sensor and the robot flange to enable vision-based robotic tasks. However, this process typically requires multiple movements of the robot arm and an external calibration object, making it both time-consuming and inconvenient, especially in scenarios where frequent recalibration is necessary. In this work, we extend our previous method which eliminates the need for external calibration objects such as a chessboard. We propose a generic dataset generation approach for point cloud registration, focusing on aligning the robot base point cloud with the scanned data. Furthermore, a more detailed simulation study is conducted involving several different collaborative robot arms, followed by real-world experiments in an industrial setting. Our improved method is simulated and evaluated using a total of 14 robotic arms from 9 different brands, including KUKA, Universal Robots, UFACTORY, and Franka Emika, all of which are widely used in the field of collaborative robotics. Physical experiments demonstrate that our extended approach achieves performance comparable to existing commercial hand-eye calibration solutions, while completing the entire calibration procedure in just a few seconds. In addition, we provide a user-friendly hand-eye calibration solution, with the code publicly available at \url{github.com/leihui6/LRBO}.
\end{abstract}


\section{Introduction}

% 第一段
% point cloud data's advantages over 2D systems
% importance and applications of sensor-robot integration
% core challenges and importance of hand-eye coordination
% limitations of current methods

\begin{figure}[htbp]
\centering
\includegraphics[width=0.9\linewidth]{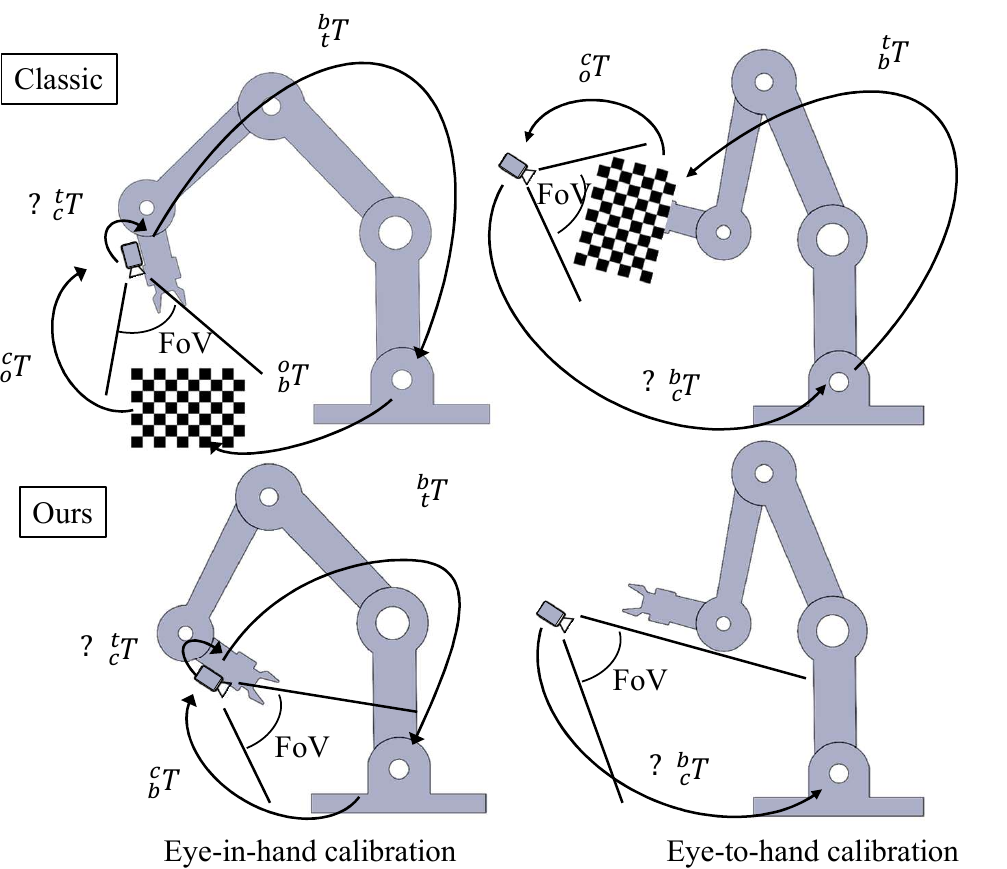}
\caption{Our proposed hand-eye and the traditional calibration that needs the external calibration objects.}
\label{fig.method}
\vspace{-4mm}
\end{figure}

Three-dimensional vision systems, particularly those utilizing point cloud data, provide detailed geometric surface information that enables comprehensive spatial understanding \cite{munaro20163d}. This capability has significantly advanced modern manufacturing systems \cite{wang2021trajectory, wang2017ubiquitous}, especially in the fields of collaborative robot and automated operations \cite{robinson2023robotic,halme2018review}. Furthermore, the integration of perception sensors with robotic manipulators has greatly enhanced automation applications that require vision-based intelligent control and manipulation \cite{zhou2021deep, ten2017grasp, zhou2021path}. A fundamental problem in collaborative robot is establishing accurate coordination between the sensing system (eye) and the tool center point (hand) \cite{horaud1995hand, jiang2022overview, enebuse2021comparative}. This coordination, known as hand-eye calibration, determines the spatial transformation, which consists of both translation and rotation. Widely used methodologies \cite{enebuse2021comparative, enebuse2022accuracy, wijesoma1993eye} typically depend on specialized calibration equipment and complex procedures, creating practical limitations in dynamic industrial settings where systems require frequent recalibration. In this paper, we aim to achieve fast and effective hand-eye calibration through 3D vision.

% ? the mandatory manual integration of calibration targets,

% 第二段
% traditional calibration requirements
% conventional calibration process
% limitations of conventional methods
% scenarios where limitations become more problematic

Calibration targets in conventional hand-eye calibration are essential components \cite{tsai1989new, strobl2006optimal}, with checkerboards and circles being the most commonly used standard patterns \cite{mallon2007pattern}. Through executing multiple sets of collaborative robot arm movements while simultaneously acquiring corresponding images, the homogeneous transform equation is constructed and solved to estimate the hand-eye transformation \cite{wang2024one}. Such conventional target-based approaches suffer from several inherent limitations: the mandatory manual integration of calibration targets, the inevitably approximate solutions due to its nonconvex nature \cite{wu2020globally}, and the multiple robot movements required for data collection result in significant time consumption\cite{zhou2023simultaneously, allegro2024multi, ma2018modeling}. 

% In industrial scenarios, the workload associated with conventional hand-eye calibration increases substantially in various contexts: multi-robot environments with diverse operational tasks \cite{zhou2023simultaneously}, multi-camera setups requiring continuous monitoring of human-robot collaboration \cite{allegro2024multi}, scenarios necessitating frequent camera position adjustments, or situations demanding recalibration due to adverse factors \cite{ma2018modeling}.

% 第三段
% proposed solution based on previous work
% technical contributions

To address the challenges in traditional hand-eye calibration and enhance it, we build upon our previous work\cite{li2024automatic}. This method preserves its user-friendly characteristics by eliminating the need for additional calibration targets and utilizing the robot base itself as a unified calibration target, enabling rapid hand-eye calibration. In this paper, a generic dataset generation method is proposed to make our approach applicable to a wide range of collaborative robots. Additionally, a total of 14 robotic arms are involved in our simulation environment, where the robustness and accuracy of our method are thoroughly evaluated, demonstrating the feasibility of our approach. Finally, a physical experiment is conducted, comparing our method with the current commercial hand-eye calibration solution, demonstrating its applicability in realistic use.
% robot base registration-specified deep learning network is proposed to align the acquired point cloud with its corresponding 3D model representation. This directly accomplishes hand-eye calibration through the determination of either the hand-eye transformation matrix or eye-hand transformation matrix. Additionally, we propose a model-free approach that leverages a learning-based framework to estimate the original frame system of the robot base, thereby directly estimating the transformation matrix without reliance on 3D models.

The key contributions of our work are summarized as follows:

\begin{enumerate}

\item We improve and develop a generic dataset generation method: in addition to capturing point clouds around the robot base on hemispheres with varying radii, we also simulate realistic robot poses by exploring a wide range of realistic joint angle combinations.

\item A total of 14 collaborative robot arms from 9 different brands are involved in our study, with each robot arm contributing 900 hand-eye calibration results. Both individual and overall results are analyzed, demonstrating the generalizability of our approach.

\item Real-world experiments are conducted in an industrial setting, where our proposed method is compared with a commercial calibration setup. Our method is validated using both single and multiple joint configurations, demonstrating its effectiveness and comparable performance.

\end{enumerate}

% 需要修改
% 需要修改
% 需要修改

% The remaining parts of this paper are organized as follows: In Section 2, we review related work and developments in hand-eye calibration. In Section 3, we present our deep learning network approach for hand-eye calibration of robotic arm systems. In Section 4, we describe our model-free approach for end-to-end hand-eye calibration. In Section 5, we evaluate the robustness and accuracy of our method through ground-truth-based experiments. In Section 6, we summarize the main conclusions and discuss future work.

\section{Related Work}

Hand-eye calibration is a prerequisite for vision-guided robotic manipulation systems using collaborative robot arms \cite{hong2025generative}, and the accuracy of this calibration fundamentally determines the subsequent precision of vision-based control and manipulation within the system. To compute the unknown transformation between a robot end-effector and a camera, the methodological approaches to addressing hand-eye calibration can be categorized into \cite{enebuse2021comparative}: solving homogeneous transform equations, reprojection error minimization techniques, and learning-based methods.

The conventional equation formulations can be represented as \cite{zhu2024point}:$A X = X B$ \cite{shiu1987calibration} or $A X = Y B$ \cite{zhuang1994simultaneous}. The former approach determines the hand-eye matrix $X$ through multiple sets of robot movements. To eliminate error propagation inherent in this formulation and improve noise sensitivity, Zhuang and Shiu \cite{zuang1993noise} proposed a sequential method for solving matrix $X$. Recently, Ma et al. \cite{ma2016new} proposed novel probabilistic approaches that extend the Batch method to filter outlier data. The second equation seeks to simultaneously determine both the hand-eye matrix $X$ and the transformation matrix $Y$ from the robot base to the world coordinate. To solve this equation, many solution strategies utilize linear least-squares minimization combined with iterative optimization schemes \cite{tabb2017solving}. In response to these approaches, Ha \cite{ha2022probabilistic} proposed a probabilistic framework that elucidates the ambiguous aspects of existing methods by revealing their underlying assumptions about system noise. An alternative approach leverages reprojection error minimization for hand-eye calibration. Based on this principle, Kenji and Emanuele \cite{koide2019general} proposed a methodology that accommodates diverse camera models through projection model adaptation, providing an efficient and robust solution to the estimation problem. However, these conventional methods and their derivative approaches typically require either multiple robot movements (at least two), additional calibration targets, or suffer from high algorithmic complexity and extended computational time.

To overcome the limitations of conventional approaches, learning-based methodologies have been progressively incorporated into hand-eye calibration procedures. Hua and Zeng \cite{hua2021hand} established coordinate transformation relationships through neural network training, achieving enhanced grasping accuracy even under the influence of noise perturbations. In constrained surgical scenarios, Krittin et al. \cite{pachtrachai2021learning} estimated hand-eye transformation through a deep convolutional network, leveraging temporal information between frames and kinematic data without requiring calibration objects. Bahadir et al. \cite{bahadir2024continual} proposed Continual Learning-based approaches that enable progressive extension of the calibration space without complete retraining, while accommodating changes in camera pose over time. Nevertheless, existing learning-based methods still present considerable room for improvement. The associated neural networks exhibit high computational complexity, limited generalizability for straightforward implementation in common robotic manipulators, and in some cases, continued dependence on external  calibration objects. Furthermore, certain approaches are restricted to either eye-in-hand or eye-to-hand calibration exclusively, failing to meet the requirements of both scenarios.

In comparison with the aforementioned methodologies, our proposed method maintains the advantage of calibration-object-free operation while accommodating both eye-in-hand and eye-to-hand calibration configurations. By looking at the robot base, a closed kinematic chain is established, allowing the transformation matrix between the camera and the robot flange to be determined straightforwardly. In addition, we introduce a dataset generation method for point cloud registration of the robot base, enhancing the generality of our approach. Simulations involving 14 commonly used collaborative robot arms, along with physical experiments comparing our method to commercial solutions, demonstrate the applicability and effectiveness of the proposed method.

\section{Problem Definition and Methods}

\subsection{Problem Definition}

The essence of the hand-eye calibration problem in a vision-guided collaborative robot system is to determine the transformation matrix between the camera and the tool center point (eye-in-hand calibration), or between the camera and the robot base (eye-to-hand calibration). In eye-in-hand calibration, the camera is mounted on the robot's end-effector, moving with the arm. In contrast, in eye-to-hand calibration, the camera is fixed in the environment to observe the robot's workspace. To formalize this, we define the coordinate systems: the robot base as $F_b$, the tool center point as $F_t$, calibration object as $F_o$, and the camera as $F_c$. 

\subsection{Eye-in-Hand Calibration}

In the eye-in-hand configuration, the camera is rigidly mounted on the robot’s flange, as illustrated in the first one in Figure \ref{fig.method}. The primary objective is to compute the rigid transformation matrix between the camera coordinate system and the tool center point (TCP) frame. This problem is commonly formulated as $AX=XB$, where $X$ represents the transformation to be solved. This calibration typically requires the robot to move to multiple poses while observing a calibration target. Specifically, it can be expressed as:

% \begin{equation}
% ^{world}T_{ee} \;\; ^{ee}X_{camera} =  ^{world}T_{camera}
% \end{equation}

%  \begin{equation}
% \underbrace{{}^{{TCP}_i}_{{{TCP}_j}}\mathbf{T}}_{A} \underbrace{{}^{TCP}_{Cam}\mathbf{T}}_{X} = \underbrace{{}^{TCP}_{Cam}\mathbf{T}}_{X} \underbrace{{}^{{Obj}_i}_{{{Obj}_j}}\mathbf{T}}_{B}
% \end{equation}

\begin{equation}
{}^{b}_{t}\mathbf{T}^{-1}_{i} \,{}^{b}_{t}\mathbf{T}_{j} {}^{t}_{c}\mathbf{T} = {}^{t}_{c}\mathbf{T} {}^{c}_{o}\mathbf{T}_{i} \,{}^{c}_{o}\mathbf{T}^{-1}_{j}
% \underbrace{{}^{b}_{t}\mathbf{T}^{-1}_{i} \,{}^{b}_{t}\mathbf{T}_{j}}_{A} \underbrace{{}^{t}_{c}\mathbf{T}}_{X} = \underbrace{{}^{t}_{c}\mathbf{T}}_{X} \underbrace{{}^{c}_{o}\mathbf{T}_{i} \,{}^{c}_{o}\mathbf{T}^{-1}_{j}}_{B}
\end{equation}

\noindent where indices $i$ and $j$ denote two distinct robot poses from the set of calibration measurements, and ${}^{b}_{t}\mathbf{T}$ represents the transformation matrix from the frame of the tool center point to the robot base coordinate system. ${}^{b}_{t}\mathbf{T}$ can be computed through forward kinematics, ${}^{t}_{c}\mathbf{T}$ is the unknown hand-eye transformation matrix to be determined, and ${}^{c}_{o}\mathbf{T}$ represents the transformation matrix from the calibration object coordinate system as observed by the camera to the camera coordinate system.

\subsection{Eye-to-hand Calibration}

In the eye-to-hand configuration, the camera is mounted at a fixed position within the workspace, external to the robot arm. This setup allows the camera to maintain a static, global viewpoint while the robot manipulator performs its movements. The primary objective of eye-to-hand calibration is to determine the rigid transformation matrix between the camera coordinate system and the robot base coordinate system. Through multiple movements of the calibration board at the robot arm's end-effector, the mathematical relationship can be established:

\begin{equation}
{}^{t}_{b}\mathbf{T}^{-1}_{i} \,{}^{t}_{b}\mathbf{T}_{j} {}^{b}_{c}\mathbf{T} = {}^{b}_{c}\mathbf{T} {}^{c}_{o}\mathbf{T}_{i} \,{}^{c}_{o}\mathbf{T}^{-1}_{j}
% \underbrace{{}^{t}_{b}\mathbf{T}^{-1}_{i} \,{}^{t}_{b}\mathbf{T}_{j}}_{A} \underbrace{{}^{b}_{c}\mathbf{T}}_{X} = \underbrace{{}^{b}_{c}\mathbf{T}}_{X} \underbrace{{}^{c}_{o}\mathbf{T}_{i} \,{}^{c}_{o}\mathbf{T}^{-1}_{j}}_{B}
\end{equation}

\noindent  where ${}^{b}_{c}\mathbf{T}$ is the unknown transformation from the camera coordinate system to the robot base coordinate system to be calculated, and ${}^{c}_{o}\mathbf{T}$ represents the transformation from the camera coordinate system to the observed calibration target measured at different robot poses.

\subsection{Look at Robot Base Once}

The proposed methodology eliminates the need for a dedicated calibration object by directly observing the robot base, as illustrated in the second setup in Figure~\ref{fig.method}. Given the 3D data of the robot base, its 6D pose can be estimated, allowing the determination of the transformation matrix between the camera coordinate system and the robot base coordinate system. This approach can be further extended to compute the transformation matrix between the TCP coordinate system and the camera coordinate system.

Given the captured robot base point cloud $P = \{p_i \in \mathbb{R}^3 | i = 1 \ldots n\}$ and the reference model point cloud $Q = \{q_j \in \mathbb{R}^3 | j = 1 \ldots m\}$, the objective of registration is to find the optimal rigid body transformation parameters: an orthogonal rotation matrix $R \in \text{SO}(3)$ and translation vector $t \in \mathbb{R}^3$ that minimize the error between the transformed actual point cloud and the reference model point cloud. The optimization objective can be expressed as: 
\begin{equation}
\min_{R,t} \sum_{i=1}^{n} \min_{j=1}^{m} \|p_i - (Rq_j + t)\|_2
\end{equation}
\noindent where $\|p_i - (Rq_j + t)\|_2$ is the Euclidean distance between the acquired point cloud and the reference point cloud of robot base. The reference model and the acquired data differ in scale; therefore, a pre-transformation of the reference data, denoted as ${}^{{ref}^{\prime}}_{ref}\mathbf{T}$, is necessary. The registration module then provides the transformation ${}^{c}_{{ref}^{\prime}}\mathbf{T}$, leading to the formulation in Eq. \ref{eq.1}.
\begin{align}
\label{eq.1}
    {}^{c}_{ref}\mathbf{T} &= {}^{c}_{{ref}^{\prime}}\mathbf{T} \, \, \,\, {}^{{ref}^{\prime}}_{ref}\mathbf{T} \\
\label{eq.2}
    {}^{b}_{c}\mathbf{T} &= {}^{c}_{ref}\mathbf{T}^{-1}
\end{align}
Here, the frame of the reference data is aligned with the frame of the robot base in the real world, which gives Eq. \ref{eq.2}. For eye-in-hand calibration, a closed kinematic chain can be constructed for each movement as
\begin{equation}
    \mathbf{I} = {}^{b}_{t}\mathbf{T} \,\, {}^{t}_{c}\mathbf{T} \,\, {}^{c}_{b}\mathbf{T}
\end{equation}
The transformation matrix ${}^{t}_{c}\mathbf{T}$ therefore can be obtained by performing the transformation as
\begin{equation}
{}^{t}_{c}\mathbf{T} = {}^{b}_{t}\mathbf{T}^{-1} \,\, {}^{c}_{b}\mathbf{T}^{-1}
\end{equation}

% The obtained rotation matrix $R$ and translation vector $t$ can be used to construct the transformation matrix ${}^{c}_{{ref}^{\prime}}\mathbf{T}$.

% If the coordinate system of the real world is identical to the $F_{ref}$ coordinate system, then $F_{ref}$ and $F_b$ are aligned. In the case of the eye-in-hand configuration, the solution for ${}^{b}_{c}\mathbf{T}$, as shown in the following equation:

% \begin{equation}
%     {}^{b}_{c}\mathbf{T} =  {}^{c}_{ref}\mathbf{T}^{-1}
% \end{equation}
The eye-in-hand setup is more computationally challenging due to the moving camera frame, whereas in eye-to-hand calibration, ${}^{b}_{c}\mathbf{T}$ can be estimated straightforwardly using a point cloud registration module. Therefore, we adopt the eye-in-hand configuration for experimental validation to demonstrate the robustness of our approach.

% After obtaining the target transformation matrix, there will be errors introduced by registration network, as well as errors between different sources in the real world affecting the final result. 

% While this mathematical formulation provides a theoretical solution, practical implementations face challenges due to various error sources. The registration process introduces errors from framework, sensor noise, discretization, and algorithmic approximations. Additionally, physical discrepancies between the idealized model and the real-world system contribute to inaccuracies in the final calibration. To achieve higher precision hand-eye calibration results, the results from multiple frames are input into a well-constructed deep learning framework, effectively reducing the influence of noise from different sources on the computation, and producing high-precision calibration results.

The main challenge lies in generating a dataset of robot base point clouds for registration training, where both the source data (captured point clouds) and the target data (reference model) are represented as point clouds. We begin by capturing the robot base through simulated 3D camera views positioned around a sampled hemisphere centered at the base origin. Additionally, to enhance realism, we augment the dataset by simulating valid robot poses based on joint constraints, reflecting realistic configurations seen in actual applications. Both datasets are used for training the point cloud registration network. 

\begin{figure}[htbp]
\centering
\includegraphics[width=1.0\linewidth]{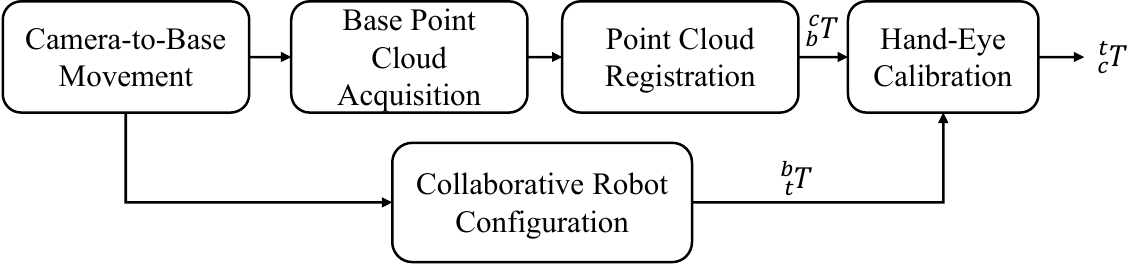}
\caption{Flowchart of the our developed hand-eye calibration.}
\label{fig.flowchat}
\end{figure}

The flowchart of our proposed method is illustrated in Figure \ref{fig.flowchat}, where the first step involves moving the robot arm to position the robot base within the camera's field of view. Given the $^{c}_{b}\mathbf{T}$, which can be estimated by the point cloud registration module, and $^{b}_{t}\mathbf{T}$, provided by forward kinematics, the $^{t}_{c}\mathbf{T}$ can finally be estimated as the result of the hand-eye calibration.

\section{Experiments}

\subsection{Preparation and Experiment Setup}
The robot arms used in our simulation environment are listed in Table \ref{tab.all_robot}. We utilize PyBullet as the simulation platform, where the robot arm is represented in URDF and OBJ file formats for loading and generating the trainable dataset. PREDATOR \cite{huang2021predator}, a registration network designed for low-overlap point clouds, is adopted in our study. The registration framework was trained on a system running Ubuntu 24.04, equipped with an Intel i9 processor and an NVIDIA RTX 3090 GPU.

In this study, the camera is positioned on a virtual hemisphere with radii of 0.5 and 0.7 meters, capturing data from multiple perspectives to simulate real 3D camera acquisition. The final dataset, available online at \footnote{huggingface.co/datasets/Leihui/lrbo}, which is combined with realistic simulated acquisition data, is used for training and is detailed in Table \ref{tab.dataset}. Five random transformation matrices are applied to the captured data. In addition, the point clouds are downsampled by a voxel size of 2 mm.

\begin{table}[htbp]
\centering
\caption{The size of the dataset for registration training.}
\label{tab.dataset}
\resizebox{0.8\columnwidth}{!}{%
\begin{tabular}{@{}ccccccc@{}}
\toprule
\textbf{FR3} & \textbf{Gen3} & \textbf{iiwa 7} & \textbf{iiwa 14} & \textbf{\begin{tabular}[c]{@{}c@{}}CRX\\ -10iA\end{tabular}} & \textbf{\begin{tabular}[c]{@{}c@{}}CRX\\ -5iA\end{tabular}} & \textbf{UR10e} \\ \midrule
1224 & 1074 & 810 & 954 & 1230 & 1026 & 816 \\ \midrule
\textbf{UR5e} & \textbf{xArm6} & \textbf{xArm7} & \textbf{\begin{tabular}[c]{@{}c@{}}CRB \\ 15000\end{tabular}} & \textbf{HC10} & \textbf{EC66} & \textbf{EC63} \\ \midrule
1110 & 756 & 1140 & 768 & 1104 & 1014 & 1140 \\ \bottomrule
\end{tabular}%
}
\end{table}

\begin{table}[htbp]
\centering
\caption{The collaborative robot arm used in our study.}
\tiny
\label{tab.all_robot}
\resizebox{0.8\columnwidth}{!}{%
\begin{tabular}{@{}lll@{}}
\toprule
\multicolumn{1}{c}{\textbf{Index}} & \multicolumn{1}{c}{\textbf{Robot Arm}} & \multicolumn{1}{c}{\textbf{Manufacturer}} \\ \midrule
1  & FR3            & Franka Robotics  \\
2  & Gen3           & Kinova           \\
3  & iiwa 7         & KUKA             \\
4  & iiwa 14        & KUKA             \\
5 & CRX-5iA        & FANUC              \\
6  & CRX-10iA       & FANUC            \\
7  & UR10e          & Universal Robots \\
8  & UR5e           & Universal Robots \\
9  & xArm6          & UFACTORY         \\
10  & xArm7          & UFACTORY         \\
11 & CRB 15000      & ABB              \\
12 & HC10           & Yaskawa Motoman  \\
13 & EC66           & Elite Robots     \\
14 & EC63           & Elite Robots     \\ \bottomrule
\end{tabular}%
}
\vspace{-2mm}
\end{table}

\subsection{Experimental Results}
\begin{figure*}[htbp]
    \centering
    \includegraphics[width=1.0\linewidth]{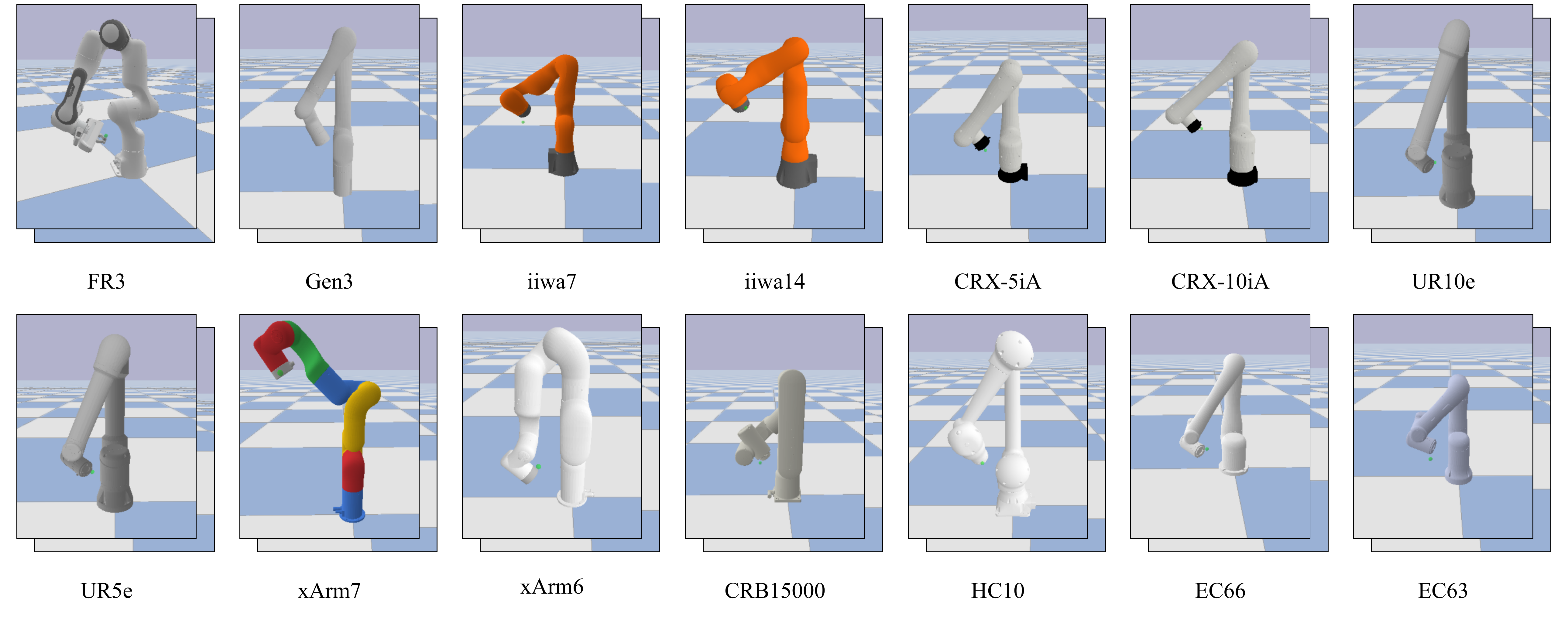}
    \caption{Visualization of randomly selected robot base-looking poses for various robotic manipulators, with one representative joint configuration shown.}
    \label{fig.14arms}
\end{figure*}

We evaluate our proposed approach in a simulation environment where the ground-truth transformation matrix between the camera and the TCP frame is known. For each robot arm, 30 random poses are selected where the camera looks at the robot base, with joint rotations of at least 20 degrees to reflect real-world conditions, as shown in Figure. \ref{fig.14arms}. At each pose, the 3D camera on the end-effector captures multiple scans of the base. Notably, each scan yields an individual calibration result. 
The final calibration result is obtained by averaging the rotation matrices in quaternion space and the translation vectors in Euclidean space. Ideally, the hand-eye calibration results obtained from the poses match the ground truth ($\bar{\mathbf{R}} \; \text{and} \; \bar{\mathbf{t}}$), meaning the RTE and RRE in translation and rotation should be zero, as %The deviation from the ground truth pose is measured as follows:
\begin{align}
    &\text{RTE} = \| \mathbf{t} - \bar{\mathbf{t}} \|_2 \\
    &\text{RRE} = \arccos \left( \frac{\text{trace}(\mathbf{R}^T \bar{\mathbf{R}}) - 1}{2} \right)
\end{align}
% where \textbf{R} and \textbf{t} denote the estimated and averaged rotation matrix and translation vector, respectively. The rotation and position deviation calculated in the simulation environment are provided in Figure. \ref{fig.sim_res}, where 5, 30 and 60 scans of the robot base are given.

% \begin{figure}[htbp]
% \centering
% \includegraphics[width=1.0\linewidth]{photo/offsets_sim.pdf}
% \caption{The result of hand-eye calibration calculated in the simulation environment.}
% \label{fig.sim_res}
% \end{figure}

\begin{figure}[htbp]
\centering
\includegraphics[width=0.9\linewidth]{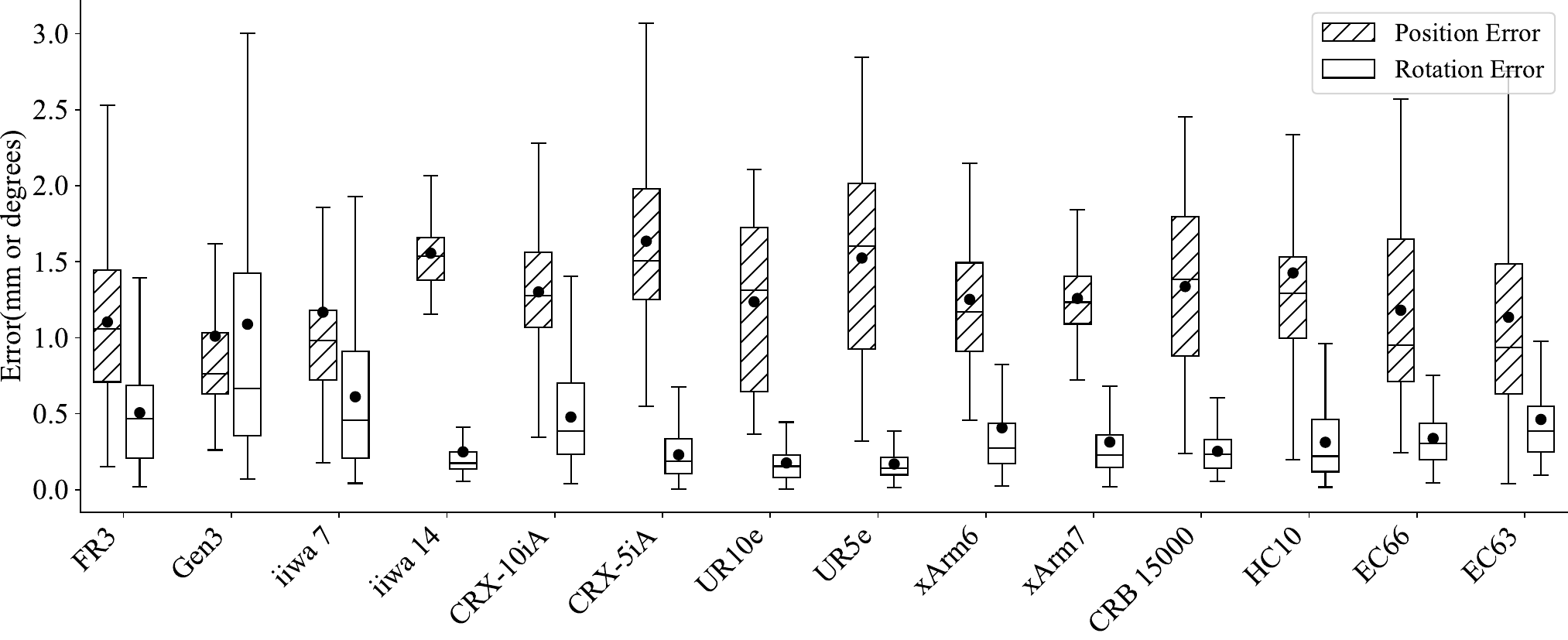}
\caption{The hand-eye calibration results calculated from all poses for each robot arm are shown, with the mean value represented by a black dot. In the box plot, the horizontal line indicates the median value, while the top and bottom edges represent the standard deviation.}
\label{fig.900_res}
\end{figure}

% As shown in Figure \ref{fig.sim_res}, the average rotation deviation is less than 0.5 degrees, and the position deviation is under 1 mm with only 5 scans. Furthermore, the performance achieved with 30 scans is nearly identical to that with 60 scans, suggesting that 30 scans of the robot base are sufficient to significantly improve calibration accuracy while additional scans do not provide further benefits. 
In addition to 30 unique base-looking poses for each robotic arm, we captured 30 repeated scans per pose, resulting in 900 calibration results per robot. The calibration performance across all robot arms is analyzed. As shown in Figure \ref{fig.900_res}, it illustrates the average deviations in position and rotation errors, and reports the median and standard deviation for each robot. According to the results, the positional deviation for all poses remains below approximately 1.5 mm, and most rotational deviations are under 0.5$^{\circ}$. The largest deviation is observed in the Gen3 robot arm. Overall, considering the point cloud voxel size of 2 mm, the tested poses yield reliable calibration performance, with a mean error of 1.29 mm in position and 0.39$^{\circ}$ in rotation.

\begin{figure}[htbp]
\centering
\includegraphics[width=1.0\linewidth]{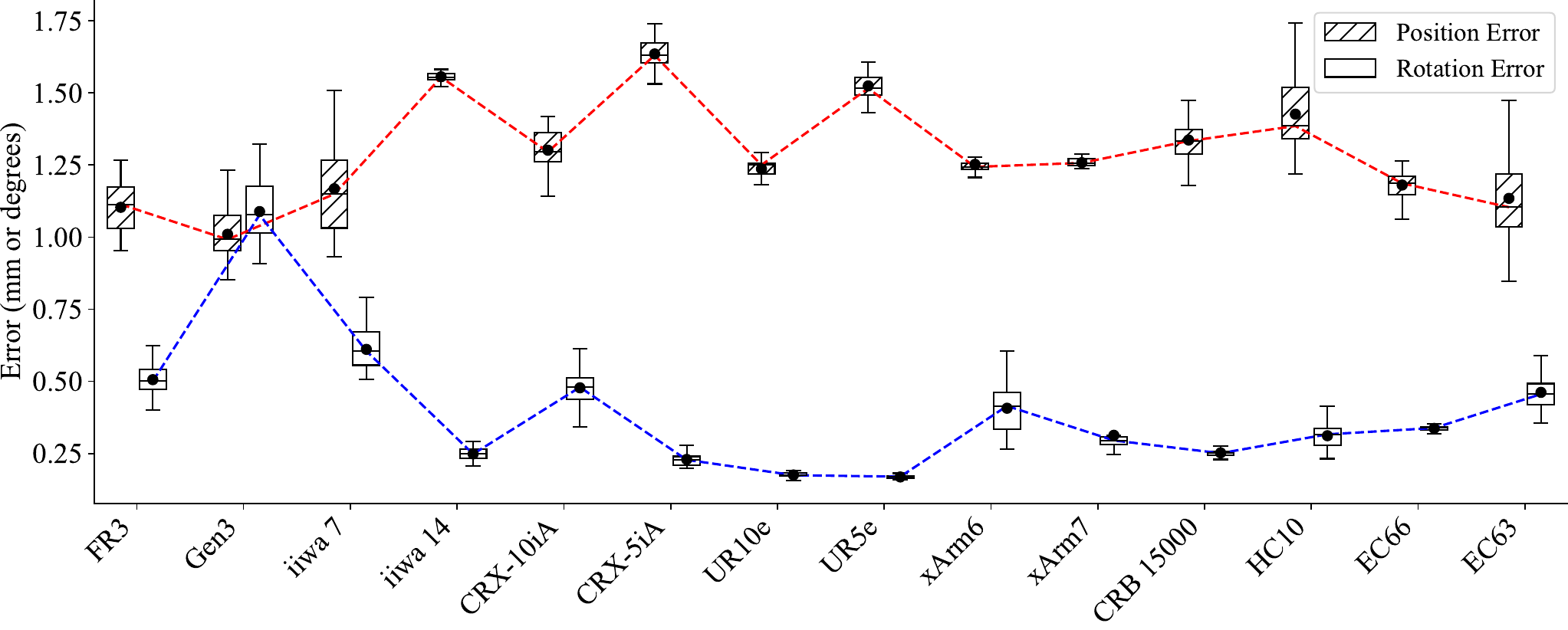}
\caption{The differences between poses are illustrated by data points connected with red and blue lines, representing positional and rotational deviations, respectively.}
\label{fig.each_pose}
\vspace{-4mm}
\end{figure}

At each pose, 30 sets of 3D data from the robot base are collected, resulting in 30 calibration results. The deviation of different poses is shown in Figure. \ref{fig.each_pose}. According to the results, the positional deviation for each pose ranges from 1 mm to 1.6 mm, with the largest variation observed in robot CRX-5iA. The rotational deviation ranges from 0.16$^{\circ}$ to 1$^{\circ}$, with the highest occurring in the Gen3 robot. The results indicate that, for the tested robot arm with different poses, the proposed method generally achieves a similar level of accuracy; that is, our method does not rely on any specific pose of the robot arm.

This deviation, observed in the simulation environment, primarily arises from registration errors and the density of the point clouds, including both the retrieved data and the standard model. However, in real-world applications, additional factors such as camera imaging inaccuracies, point cloud noise, and discrepancies between the CAD model and the scanned robot base can further contribute to calibration errors. Therefore, real-world experiments are conducted to compare our proposed method with current commercial and mature solutions.

\subsection{Real-world experiments}

\begin{figure}[htbp]
\centering
\includegraphics[width=0.7\linewidth]{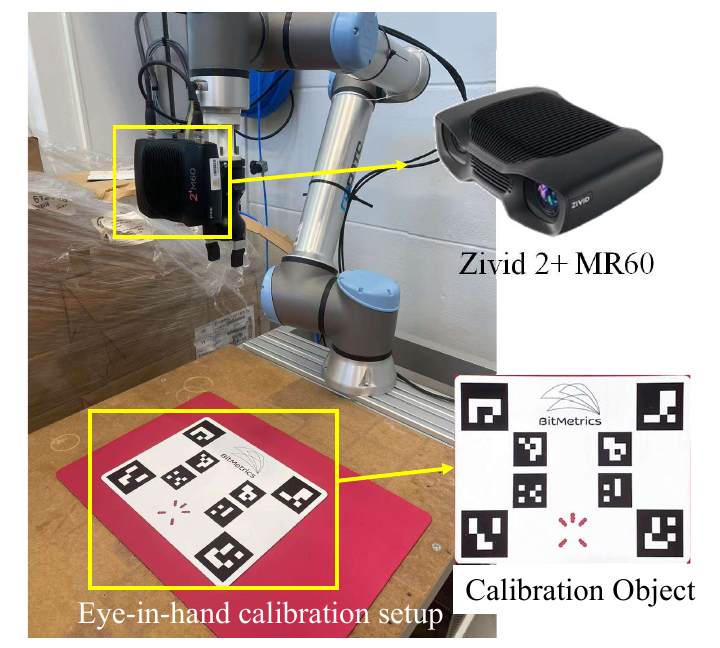}
\caption{The traditional calibration where the calibration object and multiple movement of robot arm needed.}
\label{fig.tra_exp}
\vspace{-4mm}
\end{figure}

\begin{figure}[htbp]
\centering
\includegraphics[width=0.7\linewidth]{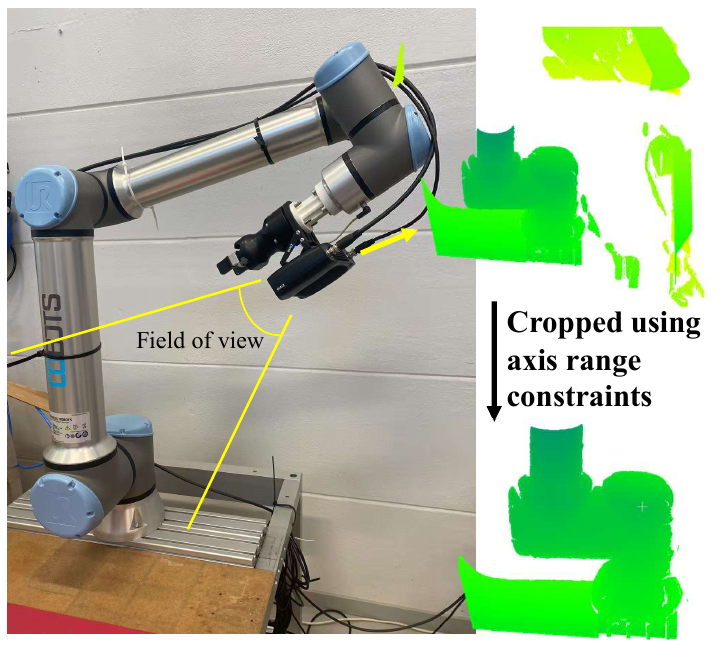}
\caption{Setup of our calibration, where no external calibration object is needed and the focus is on the robot base.}
\label{fig.our_exp}
\end{figure}

The experiment is conducted on the UR10e robot with a high-accuracy 3D camera, the Zivid 2+ MR60, which achieves a spatial resolution of 0.24 mm at a working distance of 60 cm and generates highly detailed point clouds with a density of 5,000 points per cm². The experimental setup is shown in Figure. \ref{fig.tra_exp}, where a calibration board is needed. Figure. \ref{fig.our_exp} illustrates the calibration setup, with the camera oriented toward the robot base.

\begin{figure}[htbp]
\centering
\includegraphics[width=0.9\linewidth]{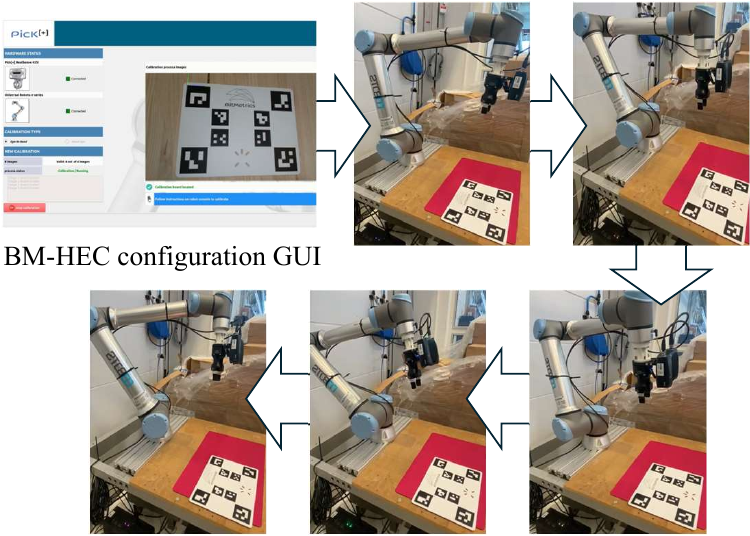}
\caption{Traditional calibration requires multiple robot movements to capture the calibration object from different perspectives.}
\label{fig.tra_exp_gui}
\end{figure}

We compared our method with the current commercial eye-in-hand calibration solution, BM-HEC\footnote{\url{www.bitmetrics.es}}, which is based on the AX = XB formulation. The traditional calibration process was repeated four times, as shown in Figure \ref{fig.tra_exp_gui}, where multiple robot movements are required during each calibration. Our method was also performed four times, with each individual scan producing a complete calibration result. The calibration process typically takes an average of 2 minutes and 48 seconds, requiring 14 poses and images of the calibration board placed on the table. In contrast, our method completes the calibration in just a few seconds with a single pose. The average calibration results, i.e., the rigid relationship between the robot flange and the camera frame, are shown in Table \ref{tab.real_res}, where the rotation in Euler space and position in Euclidean space are provided.

\begin{table}[htbp]
\centering
\caption{Comparison of Eye-in-Hand Calibration with a Commercial Calibration Solution.}
\label{tab.real_res}
\begin{tabular}{@{}lllll@{}}
\toprule
Method            & TX (m)   & TY (m)   & TZ (m)   & Time                                                                                \\ \midrule
BM-HEC & -0.054   & -0.094   & 0.127    & \multirow{2}{*}{\begin{tabular}[c]{@{}l@{}}2m48s  / \\per calib\end{tabular}} \\
Ours              & -0.051   & -0.096   & 0.125    &                                                                                     \\ \midrule
Method            & RX (rad) & RY (rad) & RZ (rad) & Time                                                                                \\ \midrule
BM-HEC & 0.013    & 0.044    & 0.017    & \multirow{2}{*}{\begin{tabular}[c]{@{}l@{}}6s  / \\per calib\end{tabular}}    \\
Ours              & 0.014    & 0.044    & 0.019    &                                                                                     \\ \bottomrule
\end{tabular}
\end{table}
According to the results in Table~\ref{tab.real_res}, The offsets between our method and BM-HEC are 3 mm, 2 mm, and 3 mm in terms of position, and 0.001 and 0.002 in rotation along the X and Z axes, respectively. The offset along the Y-axis is nearly zero. Most importantly, the calibration time of our proposed method is within 6 seconds, whereas the traditional method takes over 2.5 minutes, requiring multiple robot movements and a physical external calibration object.

\begin{figure}[htbp]
\centering
\includegraphics[width=1.0\linewidth]{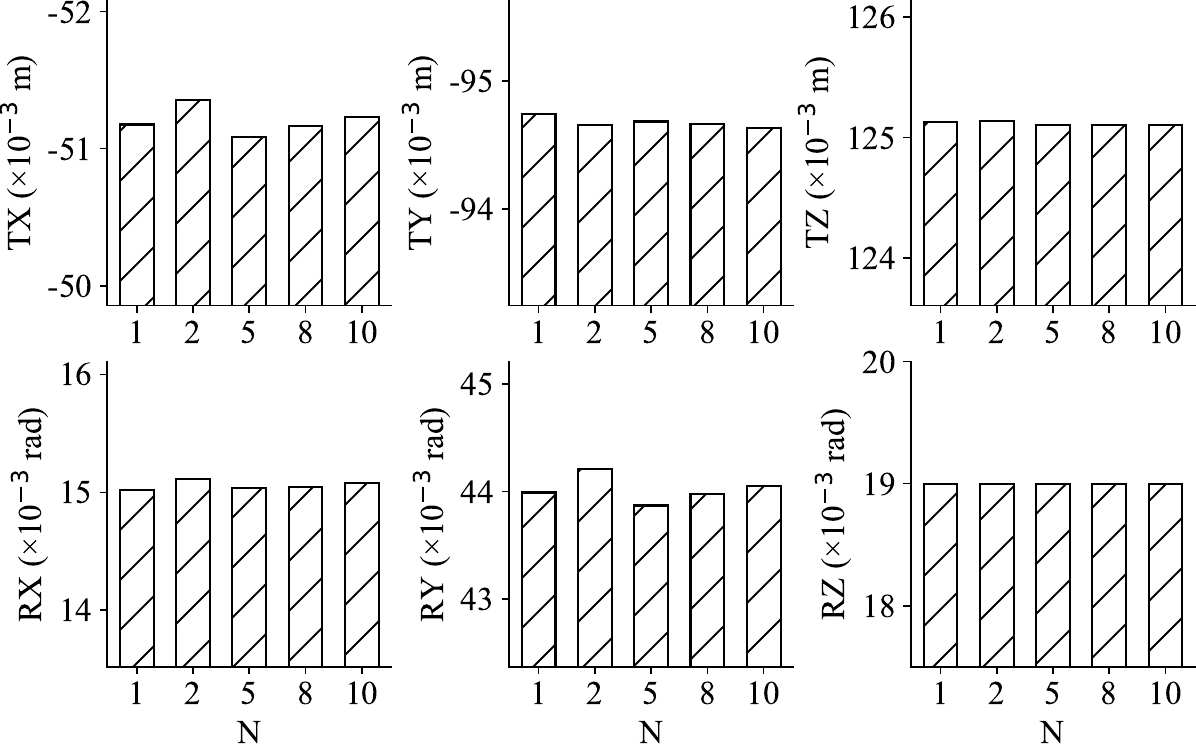}
\caption{Calibration Averaged Over the First N calibration at a Single Pose}
\label{fig.sing_pose}
\end{figure}

Furthermore, given a single pose with multiple frames captured at that pose, calibration is computed multiple times. The results are shown in Figure. \ref{fig.sing_pose}, where the average of the first N result(s) is presented for N = 1, 2, 5, 8, and 10. The results demonstrate that the largest translation deviation along the X-axis is approximately 0.2 mm, and the rotational deviation around the Y-axis is about $3 \times 10^{-4}$ radians (0.019$^\circ$). Given a standard deviation of approximately $0.1 \times 10^{-3}$, using additional frames results in nearly identical calibration performance. Therefore, our proposed method can reliably perform hand-eye calibration using just a single frame of 3D point cloud data.

\begin{figure*}[htbp]
\centering
\includegraphics[width=0.8\linewidth]{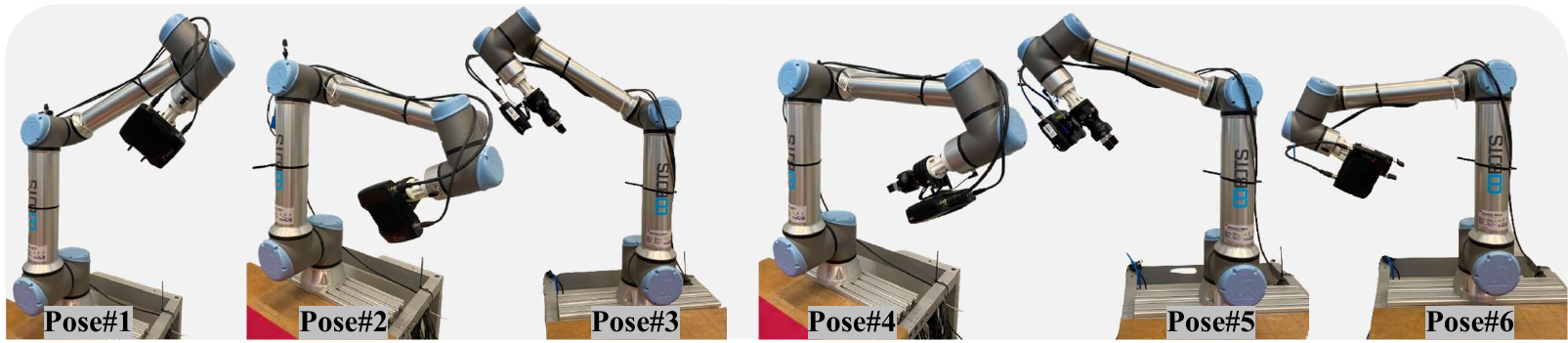}
\caption{Different poses used in our proposed hand-eye calibration method.}
\label{fig.6-pose}
\end{figure*}

\begin{figure}[htbp]
\centering
\includegraphics[width=0.9\linewidth]{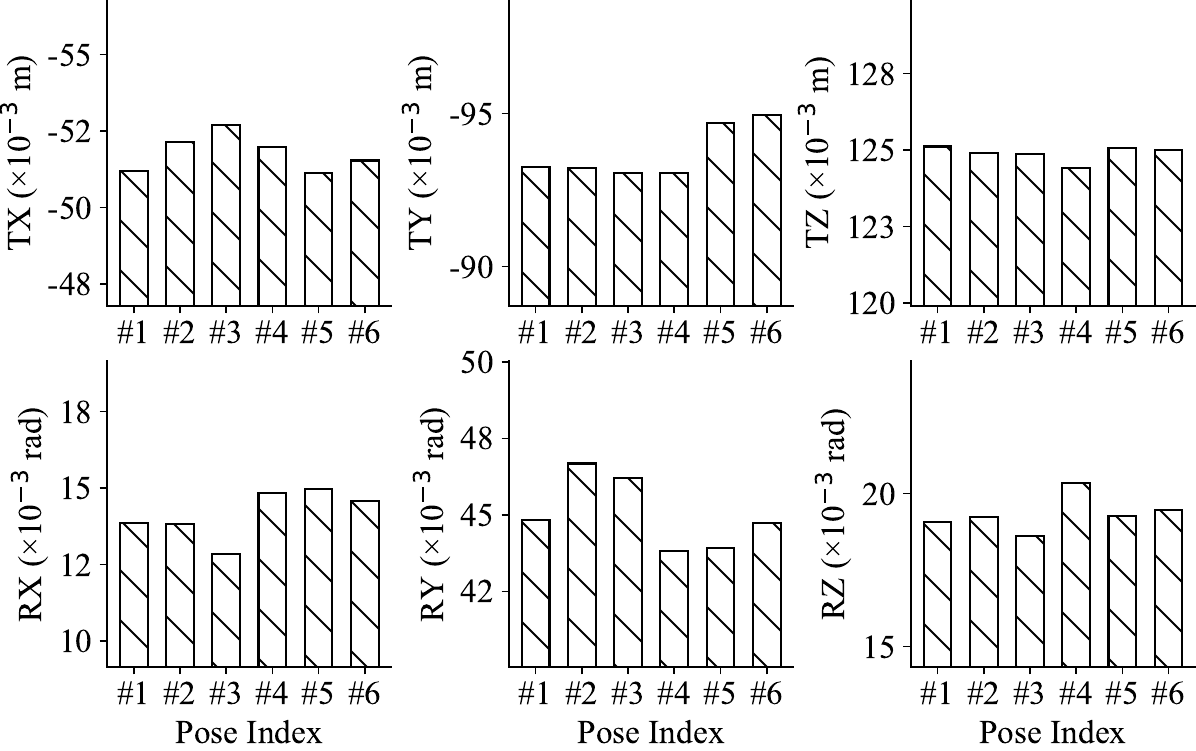}
\caption{Hand-eye calibration results across 6 poses.}
\label{fig.multi_pose}
\end{figure}

In addition, different poses for eye-in-hand calibration on the UR10e, as illustrated in Figure \ref{fig.6-pose}, are validated. The results in Figure \ref{fig.multi_pose} show translation deviations of 1.5 mm, 1.8 mm, and 0.6 mm along the X, Y, and Z axes, respectively, and rotational deviations of 0.002, 0.002, and 0.001 radians around the X, Y, and Z axes. These results demonstrate that our proposed calibration method maintains consistent performance across multiple poses, rather than depending on a specific pose.

\section{Conclusion and future work}
% In this work, we build upon our previous method, LRBO, a 3D vision-based eye-in-hand calibration approach. We conduct large-scale testing, introducing a total of 14 commonly used collaborative robots into the simulation environment. The pipeline for training robot base point cloud registration is refined and proposed. The pose of the robot base can be determined using the learning-based framework, and the transformation matrix between the robot flange and camera can be calculated within a few seconds. Meanwhile, the performance, including position and rotation errors, is comparable to existing commercial solutions.

In this work, we extended and improved our previous method, a 3D vision-based hand-eye calibration approach. Our learning-based framework estimates the pose of the robot base, enabling the computation of the transformation matrix between the robot flange and the camera within seconds. We enhance the registration dataset by incorporating captured point clouds from diverse perspectives and realistic robot arm configurations. Furthermore, we conduct large-scale evaluations involving 14 widely used collaborative robot models from 9 brands, including KUKA, ABB, and FANUC, to demonstrate the robustness and generalizability of our method. Real-world validation using high-precision 3D cameras and commonly adopted collaborative robots confirms that our approach delivers calibration performance comparable to commercial solutions, while substantially reducing calibration time and eliminating the need for external calibration objects.

In future work, we aim to apply this approach to commercial products and make it more user-friendly for workers with no prior experience. Additionally, methods to provide more stable results will be explored.

\section*{Acknowledgments}
The authors would like to thank Jialong Li for his excellent collaboration and patient support, as well as everyone in RobotLab LTH for their valuable support. The support of Jungner Company is gratefully acknowledged.

\bibliographystyle{unsrt}
\bibliography{refs}

\begin{thebibliography}{10}

\bibitem{munaro20163d}
Matteo Munaro, Radu~B Rusu, and Emanuele Menegatti.
\newblock 3d robot perception with point cloud library, 2016.

\bibitem{wang2021trajectory}
Gang Wang, Wenlong Li, Cheng Jiang, Dahu Zhu, Zhongwei Li, Wei Xu, Huan Zhao, and Han Ding.
\newblock Trajectory planning and optimization for robotic machining based on measured point cloud.
\newblock {\em IEEE transactions on robotics}, 38(3):1621--1637, 2021.

\bibitem{wang2017ubiquitous}
Xi~Vincent Wang, Lihui Wang, Abdullah Mohammed, and Mohammad Givehchi.
\newblock Ubiquitous manufacturing system based on cloud: A robotics application.
\newblock {\em Robotics and Computer-Integrated Manufacturing}, 45:116--125, 2017.

\bibitem{robinson2023robotic}
Nicole Robinson, Brendan Tidd, Dylan Campbell, Dana Kuli{\'c}, and Peter Corke.
\newblock Robotic vision for human-robot interaction and collaboration: A survey and systematic review.
\newblock {\em ACM Transactions on Human-Robot Interaction}, 12(1):1--66, 2023.

\bibitem{halme2018review}
Roni-Jussi Halme, Minna Lanz, Joni K{\"a}m{\"a}r{\"a}inen, Roel Pieters, Jyrki Latokartano, and Antti Hietanen.
\newblock Review of vision-based safety systems for human-robot collaboration.
\newblock {\em Procedia Cirp}, 72:111--116, 2018.

\bibitem{zhou2021deep}
Zhengxue Zhou, Leihui Li, Riwei Wang, and Xuping Zhang.
\newblock Deep learning on 3d object detection for automatic plug-in charging using a mobile manipulator.
\newblock In {\em 2021 IEEE International Conference on Robotics and Automation (ICRA)}, pages 4148--4154. IEEE, 2021.

\bibitem{ten2017grasp}
Andreas Ten~Pas, Marcus Gualtieri, Kate Saenko, and Robert Platt.
\newblock Grasp pose detection in point clouds.
\newblock {\em The International Journal of Robotics Research}, 36(13-14):1455--1473, 2017.

\bibitem{zhou2021path}
Peng Zhou, Rui Peng, Maggie Xu, Victor Wu, and David Navarro-Alarcon.
\newblock Path planning with automatic seam extraction over point cloud models for robotic arc welding.
\newblock {\em IEEE robotics and automation letters}, 6(3):5002--5009, 2021.

\bibitem{horaud1995hand}
Radu Horaud and Fadi Dornaika.
\newblock Hand-eye calibration.
\newblock {\em The international journal of robotics research}, 14(3):195--210, 1995.

\bibitem{jiang2022overview}
Jianfeng Jiang, Xiao Luo, Qingsheng Luo, Lijun Qiao, and Minghao Li.
\newblock An overview of hand-eye calibration.
\newblock {\em The International Journal of Advanced Manufacturing Technology}, 119(1):77--97, 2022.

\bibitem{enebuse2021comparative}
Ikenna Enebuse, Mathias Foo, Babul Salam Ksm~Kader Ibrahim, Hafiz Ahmed, Fhon Supmak, and Odongo~Steven Eyobu.
\newblock A comparative review of hand-eye calibration techniques for vision guided robots.
\newblock {\em IEEE Access}, 9:113143--113155, 2021.

\bibitem{enebuse2022accuracy}
Ikenna Enebuse, Babul KSM~Kader Ibrahim, Mathias Foo, Ranveer~S Matharu, and Hafiz Ahmed.
\newblock Accuracy evaluation of hand-eye calibration techniques for vision-guided robots.
\newblock {\em Plos one}, 17(10):e0273261, 2022.

\bibitem{wijesoma1993eye}
SW~Wijesoma, DFH Wolfe, and RJ~Richards.
\newblock Eye-to-hand coordination for vision-guided robot control applications.
\newblock {\em The International Journal of Robotics Research}, 12(1):65--78, 1993.

\bibitem{tsai1989new}
Roger~Y Tsai, Reimar~K Lenz, et~al.
\newblock A new technique for fully autonomous and efficient 3 d robotics hand/eye calibration.
\newblock {\em IEEE Transactions on robotics and automation}, 5(3):345--358, 1989.

\bibitem{strobl2006optimal}
Klaus~H Strobl and Gerd Hirzinger.
\newblock Optimal hand-eye calibration.
\newblock In {\em 2006 IEEE/RSJ international conference on intelligent robots and systems}, pages 4647--4653. IEEE, 2006.

\bibitem{mallon2007pattern}
John Mallon and Paul~F Whelan.
\newblock Which pattern? biasing aspects of planar calibration patterns and detection methods.
\newblock {\em Pattern recognition letters}, 28(8):921--930, 2007.

\bibitem{wang2024one}
Xiao Wang and Hanwen Song.
\newblock One-step solving the hand--eye calibration by dual kronecker product.
\newblock {\em Journal of Mechanisms and Robotics}, 16(10), 2024.

\bibitem{wu2020globally}
Jin Wu, Ming Liu, Yilong Zhu, Zuhao Zou, Ming-Zhe Dai, Chengxi Zhang, Yi~Jiang, and Chong Li.
\newblock Globally optimal symbolic hand-eye calibration.
\newblock {\em IEEE/ASME Transactions on Mechatronics}, 26(3):1369--1379, 2020.

\bibitem{zhou2023simultaneously}
Zishun Zhou, Liping Ma, Xilong Liu, Zhiqiang Cao, and Junzhi Yu.
\newblock Simultaneously calibration of multi hand--eye robot system based on graph.
\newblock {\em IEEE Transactions on Industrial Electronics}, 71(5):5010--5020, 2023.

\bibitem{allegro2024multi}
Davide Allegro, Matteo Terreran, and Stefano Ghidoni.
\newblock Multi-camera hand-eye calibration for human-robot collaboration in industrial robotic workcells.
\newblock {\em arXiv preprint arXiv:2406.11392}, 2024.

\bibitem{ma2018modeling}
Le~Ma, Patrick Bazzoli, Patrick~M Sammons, Robert~G Landers, and Douglas~A Bristow.
\newblock Modeling and calibration of high-order joint-dependent kinematic errors for industrial robots.
\newblock {\em Robotics and Computer-Integrated Manufacturing}, 50:153--167, 2018.

\bibitem{li2024automatic}
Leihui Li, Xingyu Yang, Riwei Wang, and Xuping Zhang.
\newblock Automatic robot hand-eye calibration enabled by learning-based 3d vision.
\newblock {\em Journal of Intelligent \& Robotic Systems}, 110(3):130, 2024.

\bibitem{hong2025generative}
Ilkwon Hong and Junhyoung Ha.
\newblock Generative adversarial networks for solving hand-eye calibration without data correspondence.
\newblock {\em IEEE Robotics and Automation Letters}, 2025.

\bibitem{zhu2024point}
Dahu Zhu, Hao Wu, Tao Ding, and Lin Hua.
\newblock Point cloud registration-enabled globally optimal hand--eye calibration.
\newblock {\em IEEE/ASME Transactions on Mechatronics}, 2024.

\bibitem{shiu1987calibration}
Yiu~Cheung Shiu and Shaheen Ahmad.
\newblock Calibration of wrist-mounted robotic sensors by solving homogeneous transform equations of the form ax= xb.
\newblock 1987.

\bibitem{zhuang1994simultaneous}
Hanqi Zhuang, Zvi~S Roth, and Raghavan Sudhakar.
\newblock Simultaneous robot/world and tool/flange calibration by solving homogeneous transformation equations of the form ax= yb.
\newblock {\em IEEE Transactions on Robotics and Automation}, 10(4):549--554, 1994.

\bibitem{zuang1993noise}
Hanqi Zuang and Yiu~Cheung Shiu.
\newblock A noise-tolerant algorithm for robotic hand-eye calibration with or without sensor orientation measurement.
\newblock {\em IEEE transactions on systems, man, and cybernetics}, 23(4):1168--1175, 1993.

\bibitem{ma2016new}
Qianli Ma, Haiyuan Li, and Gregory~S Chirikjian.
\newblock New probabilistic approaches to the ax= xb hand-eye calibration without correspondence.
\newblock In {\em 2016 IEEE international conference on robotics and automation (ICRA)}, pages 4365--4371. IEEE, 2016.

\bibitem{tabb2017solving}
Amy Tabb and Khalil~M Ahmad~Yousef.
\newblock Solving the robot-world hand-eye (s) calibration problem with iterative methods.
\newblock {\em Machine Vision and Applications}, 28(5):569--590, 2017.

\bibitem{ha2022probabilistic}
Junhyoung Ha.
\newblock Probabilistic framework for hand--eye and robot--world calibration $ ax= yb$.
\newblock {\em IEEE Transactions on Robotics}, 39(2):1196--1211, 2022.

\bibitem{koide2019general}
Kenji Koide and Emanuele Menegatti.
\newblock General hand--eye calibration based on reprojection error minimization.
\newblock {\em IEEE Robotics and Automation Letters}, 4(2):1021--1028, 2019.

\bibitem{hua2021hand}
Jiang Hua and Liangcai Zeng.
\newblock Hand--eye calibration algorithm based on an optimized neural network.
\newblock In {\em Actuators}, volume~10, page~85. MDPI, 2021.

\bibitem{pachtrachai2021learning}
Krittin Pachtrachai, Francisco Vasconcelos, Philip Edwards, and Danail Stoyanov.
\newblock Learning to calibrate-estimating the hand-eye transformation without calibration objects.
\newblock {\em IEEE Robotics and Automation Letters}, 6(4):7309--7316, 2021.

\bibitem{bahadir2024continual}
Ozan Bahadir, Jan~Paul Siebert, and Gerardo Aragon-Camarasa.
\newblock Continual learning approaches to hand--eye calibration in robots.
\newblock {\em Machine Vision and Applications}, 35(4):97, 2024.

\bibitem{huang2021predator}
Shengyu Huang, Zan Gojcic, Mikhail Usvyatsov, Andreas Wieser, and Konrad Schindler.
\newblock Predator: Registration of 3d point clouds with low overlap.
\newblock In {\em Proceedings of the IEEE/CVF Conference on computer vision and pattern recognition}, pages 4267--4276, 2021.

\end{thebibliography}

\end{document}